\documentclass[pmlr]{jmlr}

\usepackage{float}

\usepackage{longtable}

\usepackage{adjustbox}
\usepackage{booktabs}
\usepackage[load-configurations=version-1]{siunitx} 

\usepackage [english]{babel}
\usepackage [autostyle, english = american]{csquotes}
\MakeOuterQuote{"}

\makeatletter
\def\set@curr@file#1{\def\@curr@file{#1}} 
\makeatother

\theorembodyfont{\upshape}
\theoremheaderfont{\scshape}
\theorempostheader{:}
\theoremsep{\newline}

\jmlrvolume{126}
\jmlryear{2020}
\jmlrworkshop{Machine Learning for Healthcare}

\title[ Robust Benchmarking for Machine Learning of Clinical Entity Extraction]{ Robust Benchmarking for Machine Learning of\\ Clinical Entity Extraction
}

\author{\Name{Monica Agrawal}
      \Email{magrawal@mit.edu}\\ 
      \addr Department of Electrical Engineering and Computer Science\\
      Massachusetts Institute of Technology \\
      Cambridge, MA, USA
      \AND
      \Name{Chloe O'Connell}
      \Email{coconell13@partners.org}\\ 
      \addr Department of Anesthesia, Critical Care and Pain Medicine\\
      Massachusetts General Hospital \\
      Boston, MA, USA
    \AND
      \Name{Yasmin Fatemi}
      \Email{yfatemi@partners.org}\\ 
      \addr Department of Anesthesia, Critical Care and Pain Medicine\\
      Massachusetts General Hospital \\
      Boston, MA, USA
      \AND
      \Name{Ariel Levy}
      \Email{aslevy@alum.mit.edu}\\ 
      \addr Department of Electrical Engineering and Computer Science\\
      Massachusetts Institute of Technology \\
      Cambridge, MA, USA
      \AND
      \Name{David Sontag}
      \Email{dsontag@csail.mit.edu}\\ 
      \addr Department of Electrical Engineering and Computer Science\\
      Massachusetts Institute of Technology \\
      Cambridge, MA, USA
       } 

\editor{Editor's name}

\begin{document}

\maketitle

\begin{abstract}
Clinical studies often require understanding elements of a patient’s narrative that exist only in free text clinical notes. To transform notes into structured data for downstream use, these elements are commonly extracted and normalized to medical vocabularies. In this work, we audit the performance of and indicate areas of improvement for state-of-the-art systems. We find that high task accuracies for clinical entity normalization systems on the 2019 n2c2 Shared Task are misleading, and underlying performance is still brittle. Normalization accuracy is high for common concepts (95.3\%), but much lower for concepts unseen in training data (69.3\%). We demonstrate that current approaches are hindered in part by inconsistencies in medical vocabularies, limitations of existing labeling schemas, and narrow evaluation techniques. We reformulate the annotation framework for clinical entity extraction to factor in these issues to allow for robust end-to-end system benchmarking. We evaluate concordance of annotations from our new framework between two annotators and achieve a Jaccard similarity of 0.73 for entity recognition and an agreement of 0.83 for entity normalization. We propose a path forward to address the demonstrated need for the creation of a reference standard to spur method development in entity recognition and normalization.
\end{abstract}

\section{Introduction}
Free text clinical notes contain a rich narrative of a patient’s interactions with the healthcare system. However, the strengths and weaknesses of clinical notes lie in the diversity of natural language. Due to the flexibility of documentation, notes contain information omitted from the structured clinical record, but the language itself is often hard to parse into a consistent structured format. For example, \texttt{Cold} can refer to the temperature, a temperament, the viral infection, or Chronic Obstructive Lung Disease. By matter of preference, a doctor can refer to a patient with a 101$\degree$ F temperature as \texttt{running a fever}, \texttt{being febrile}, or \texttt{having pyrexia}. 

In order to transform text into a unified structured format useful for downstream applications, EHR text mining often involves recognizing spans representing concepts (named entity recognition) and mapping these spans to a common vocabulary (named entity normalization/linkage). We will refer to both steps together as clinical entity extraction. An example of this two-step process is shown below in Figure 1. Typically, in medicine, concepts are mapped to terms in the Unified Medical Language System (UMLS), each term denoted by a Concept Unique Identifier (CUI)~\citep{Bodenreider2004}. 

\begin{figure}[htbp]
  \centering 
  \includegraphics[width=6in]{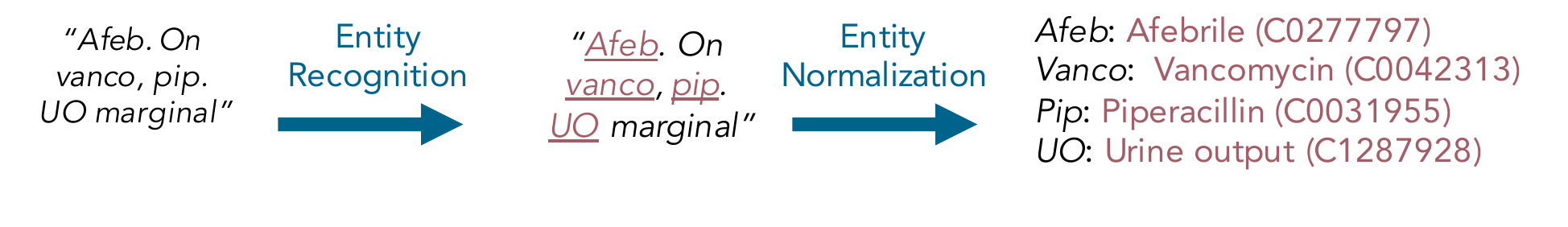} 
  \vspace{-5mm}
  \caption{Pipeline of extracting UMLS concepts from text. Typically, this is broken down into two steps as shown: recognition (detecting relevant spans) and normalization (mapping spans to a vocabulary).}
  \label{fig:example} 

\end{figure} 
Normalized clinical entities can be used as direct input for retrospective analyses and as an additional source of information when addressing confounders. For example, clinical entities can help learn adverse drug events or a health knowledge graph ~\citep{Jensen2012, Rotmensch2017, Nadkarni2011}; or at point-of-care, e.g. as a step in matching patients to relevant clinical trials ~\citep{Demner-Fushman2009}. 
Additionally, causal inference and reinforcement learning must both account for unobserved confounders affecting both medical intervention and patient outcome~\citep{Gottesman2019}, information often contained in normalized clinical entities. Structured text data could reveal relevant patient history or additional interventions omitted from the structured clinical record. Because normalizing entities condenses high-dimensional raw text into a lower-dimensional space by collapsing synonyms, relevant factors from normalized entities could be plausibly integrated into a model's state space. In a similar vein, \citet{Oakden-Rayner2020} found that machine learning models trained on medical images suffer from hidden stratification, where performance depends on unknown confounders. In particular, the authors found that models for predicting pneumothorax from X-rays were partially picking up on previously placed chest drains. The presence of these chest drains is often noted in the accompanying radiology notes. As before, finding clinical entities associated with abnormally high or low performance could identify unexpected confounders. Such multimodal sanity checks are particularly important given the high-stakes nature of machine learning for healthcare.

Despite the importance of clinical entity extraction for a range of tasks, it still remains largely an open problem, and clinical research often has to resort to manual chart review for high-fidelity data, which is tedious and difficult to scale. While there has been a significant, concerted effort from the clinical natural language processing (NLP) community to bridge this gap, performance falters due to (i) the lack of sufficient annotated training data, (ii) the huge space of clinical entities to map to, and (iii) known issues with medical vocabularies. 
\subsection*{Contributions}
In this work, we investigate, quantify performance of, and indicate areas of improvements for the current state of normalization by analyzing performance on the Medical Concept Normalization (MCN) Corpus~\citep{Luo2019}. We focus on a subset of CUIs in the UMLS: the SNOMED Clinical Terms, which consist of over 400,000 common clinical concepts; and RxNorm, a comprehensive clinical drug vocabulary~\citep{Luo2019, NIH-NLM2015, Nelson2011}.
We examine top systems for clinical entity normalization from the 2019 National NLP Clinical Challenges (n2c2) Shared Task as well as two widely used end-to-end open-source systems, cTAKES and MetaMap~\citep{Aronson2010, Savova2010}. These analyses reveal holes in performance, underscore the need for future dataset creation and method development in this space, and highlight the need for a new evaluation framework. In their description of the MCN corpus, \citet{Luo2019} recounted how their corpus annotation effort was hindered by several issues including imperfections of UMLS. Therefore, we additionally develop a new annotation framework to address these and other identified issues.  We demonstrate how our new schema adjusts to these issues and allows for more flexible evaluation of end-to-end systems.

\subsection*{Generalizable Insights about Machine Learning in the Context of Healthcare}
Due to its utility, clinical entity extraction is widely used as an initial step in many machine learning for healthcare pipelines. As the field increasingly turns to multimodal data, clinical text will become a continuously more important component of ML systems. In this work, we show that performance across normalization and end-to-end systems differ significantly on various subsets of the data. Given this observed heterogeneity in performance, it is incredibly important for users of such systems to consider how well they extract concepts relevant to their specific use cases.  Unfortunately, a full evaluation of these systems is currently unachievable since a publicly available reference standard for end-to-end benchmarking does not exist. Such reference standards have been crucial to progress in other fields in machine learning, e.g. computer vision through ImageNet~\citep{Deng2010}. In this work, we propose recommendations to the clinical NLP community regarding a path forward for the creation of that reference standard. We detail a robust annotation framework that by construction allows for end-to-end evaluation and reduces ambiguity for annotators.

\section{Related Work}

In the general domain, datasets for named entity recognition and normalization are huge, leading to the success of data-hungry algorithms. The Wiki-Disamb30 dataset, for example, consists of 1.4 million texts ~\citep{Ferragina2010}. Techniques also often pull in external information: e.g. population priors, term-term co-occurrence, rich term metadata, and knowledge graphs~\citep{Hoffart2011}. Recent successful methods include end-to-end systems and DeepType, which uses a type classifier trained on 800 million tokens~\citep{Raiman2018, Kolitsas2018}. However, these do not translate to the clinical domain because of the dearth of rich annotated clinical text. While a dataset of the magnitude seen in the general domain is infeasible in healthcare, one of our motivations for the creation of a larger dataset stems from successes in the general domain. 
Another drawback of the clinical domain is the comparative lack of entity information. UMLS contains a Related Concepts table with entity-entity relationships, including triples such as (‘Gematicin’, ‘is\_a’, ‘antibiotic’). Unfortunately, these documented relationships are unreliable and incomplete ~\citep{Halper2011, Chen2009, Agrawal2014, Cimino2003}. As a result, the table cannot be cleanly used to leverage entity-entity co-occurrences or other relationships, and more creative solutions are required.

Due to the importance of clinical entity extraction, there exist several open-sourced clinical entity extraction systems, such as cTAKES, MetaMap, and MedLEE~\citep{Savova2010, Aronson2010, Friedman2004}. These systems are very popular; over the past year, cTAKES has been mentioned in over 300 papers, MetaMap in over 500, and MedLEE in over 150. Many of these papers use these systems as a tool in their analysis. Unfortunately, past work has benchmarked the performance of these open-source systems in limited settings to middling results ~\citep{Reategui2018}; for example, ~\citet{Wu2012} found that recall was below 0.5 for all systems on abbreviations specifically. However, due to the lack of a reference dataset, such benchmark numbers don't exist over a diverse set of terms. 

Several challenges in the clinical NLP community have focused on addressing entity recognition and/or normalization~\citep{Chapman2011}. The 2010 i2b2/VA challenge released a dataset of 871 discharge summaries for recognition (but not normalization) of problems, tests, and treatments~\citep{Uzuner2011}. The 2012 i2b2 challenge released a similar dataset of 310 discharge summaries, that additionally annotated clinical departments, evidentials (e.g. ‘complained of’), and occurrences (e.g. ‘transfer’, ‘admission’). State-of-the-art for the clinical entity recognition task leverages contextual embeddings~\citep{Si2019}. However, performance there is inconsistent, e.g. with an F1 score of 85.1 for treatment span detection, but only an F1 score of 66.3 for occurence detection. 

Some clinical datasets also include span normalization. ShARe/Clef eHealth 2013 Task 1, SemEval-2014 Task 7, and SemEval-2015 Task 14 provide normalization to CUIs, but only for disease/disorder concepts~\citep{Elhadad2015, Pradhan2015, Suominen2013}. Recently, \citet{Luo2019} derived the Medical Concept Normalization Corpus (MCN) from the spans of 100 of the discharge summaries from the 2010 i2b2/VA challenge. They normalized over 13,600 entity spans (problems, tests, and treatments) to terms from SNOMED and RxNorm~\citep{Luo2019}. Track 3 of the 2019 National NLP Clinical Challenges (n2c2) Shared Tasks and Workshop utilized the MCN corpus. The challenge yielded 108 submissions, with a top submission accuracy of 85.2\%.

In their paper, the creators of the MCN corpus detailed issues they faced in normalizing terms~\citep{Luo2019}. For example, if a compound term was not available in their vocabulary, the authors split the term into its constituents instead of labeling it as CUI-less. Since there is no term for \texttt{percutaneous drains}, they instead split the term into ‘percutaneous (C0522523)’ and ‘drains (C0180499).’ Since they used the spans from the 2010 i2b2 challenge, they recounted that spans were inconsistently annotated and were suboptimal for the normalization task. They also described inherent issues in the SNOMED vocabulary, such as inconsistency, duplication, and missing terms. These issues led them to have to make self-described arbitrary decisions in normalization. In this work, we build on the findings of \citet{Luo2019} to address these issues in an annotation schema.

\section{Current Performance of Entity Normalization Systems}
 
\subsection{Data}
To understand the current state of clinical entity normalization, we analyzed the outcomes of Track 3 of the 2019 National NLP Clinical Challenges (n2c2) Shared Task, run on the MCN dataset~\citep{Luo2019}. Participants had to provide a single corresponding CUI for each provided span. In the MCN annotation process, if the original span from i2b2 (e.g. \texttt{high-grade fever}) was split up, teams were presented with the split spans (\texttt{high-grade} and \texttt{fever}), and not the original span. Teams trained on 50 tagged discharge summaries with 50 held out for evaluation. There were 108 total submissions, and here we analyze the outputs from the top 10 teams. Top teams used a variety of techniques ranging from classical NLP and information retrieval techniques to modern neural approaches, like contextual embeddings. To place performance in context of well-known systems, we compare to output from the Default Clinical Pipeline from cTAKES version 4.0.0 and the 2018 release of MetaMap~\citep{Savova2010, Aronson2010}. 

\subsection{Analysis Methodology}
We evaluate performance of the top 10 performing systems on the test data using 3 metrics: (i) highest accuracy any system achieved, (ii) average accuracy, and (iii) pooled accuracy. We define pooled accuracy as the percentage of spans that were mapped to the correct CUI by any of the top 10 systems. In our analysis, we first preprocessed the text spans to remove possessive pronouns tagged alongside the concept (e.g. ‘her diabetes’ $\rightarrow$ ‘diabetes’).  The training dataset consists of 6684 annotations and 2331 unique CUIs, but the 100 most common CUIs constitute 31.5\% of the annotations. Since the data set is heavily right-tailed, we evaluate performance for rarer terms by focusing on the following subsets of data:
\begin{itemize}
\itemsep0em
\item   All spans \textit{(All)}
\item	Multi-word spans \textit{(Multi-word)}
\item	Spans where the corresponding text is unseen in the training set \textit{(Unseen Text)}
\item	Spans where the mapped CUI is unseen in the training set \textit{(Unseen CUI)}
\item	Spans where the text isn’t a preferred name or synonym in UMLS \textit{(Not Direct Match)}
\item	Spans mapped to the 100 most common CUIs \textit{(Top 100 CUI)}
\item	Spans where the text is more commonly mapped to a different CUI \textit{(Unpopular CUI) }
\end{itemize}

\textit{Unpopular CUI} measures how well systems do on rarer CUIs, when there are other plausible options. For instance, in the training set, \texttt{pt} appears 17 times: 15 times as ‘Prothrombin time (C0033707)’, once as ‘Physical therapy (C0949766)’, and once as ‘Posterior tibial pulse (C1720310).’ A system that always defaults to ‘Prothrombin Time’ would have 88\% accuracy but no disambiguation power.  

Additionally, we investigate how accurately different concept categories, as defined by the UMLS Semantic Types, are normalized. We only consider the categories with at least 50 instances in the test set.  

Separately, we analyze the performance of cTAKES and MetaMap. These systems were not trained for this specific task or dataset but are built to provide end-to-end annotations, not just entity normalization. Therefore, to evaluate leniently, we deemed a system correct if it output any span with the correct CUI that overlapped with the test span. If the true span did not have a CUI, any output without the exact span was treated as correct. We didn’t factor in any additional spans it might have tagged.

\subsection{Analysis Results}
While overall normalization accuracy is relatively high, it is much more brittle if we look at performance on subsets of the dataset, as detailed below in Table \ref{tab:analysis}. In particular, if a text span is more commonly mapped to a different CUI, it is very rare that systems recovered the true underlying CUI. Examples everyone missed were for ‘Cold Sensation (C0234192)’ and for the physiological process of ‘Diuresis (C0445398)’ where all participants chose the term for ‘induced diuresis.’ Some fraction of errors came from choosing a concept in the wrong type hierarchy, and many are essentially pedantic, e.g. everyone mapped \texttt{tube feeds} to ‘tube feeding diet (C0311131)’ when the gold label was ‘tube feeding of patient (C0041281).’ 

\begin{table}[htbp]
  \centering 
  \caption{Performance of the top 10 systems on the entity normalization task on the MCN corpus. Accuracy is measured on different subsets of the test data and by the 3 metrics described in Evaluation Methodology 
  }
\begin{tabular}{lcccc}
\textbf{Test Subset}                  & \multicolumn{1}{l}{\textbf{\# Examples}} & \multicolumn{1}{l}{\textbf{Max Acc}} & \multicolumn{1}{l}{\textbf{Average Acc}} & \multicolumn{1}{l}{\textbf{Pooled Acc}} \\ \hline
\multicolumn{1}{l|}{All}              & \multicolumn{1}{c|}{6925}                   & \multicolumn{1}{c|}{85.2\%}               & \multicolumn{1}{c|}{81.1\%}               & 92.7\%                                       \\ \hline
\multicolumn{1}{l|}{Top 100 CUI}      & \multicolumn{1}{c|}{1891}                   & \multicolumn{1}{c|}{95.3\%}               & \multicolumn{1}{c|}{91.1\%}               & 98.1\%                                       \\ \hline
\multicolumn{1}{l|}{Multi-word}       & \multicolumn{1}{c|}{2755}                   & \multicolumn{1}{c|}{81.9\%}               & \multicolumn{1}{c|}{74.3\%}               & 91.3\%                                       \\ \hline
\multicolumn{1}{l|}{Unseen text}      & \multicolumn{1}{c|}{2910}                   & \multicolumn{1}{c|}{69.5\%}               & \multicolumn{1}{c|}{60.7\%}               & 86.0\%                                       \\ \hline
\multicolumn{1}{l|}{Unseen CUI}       & \multicolumn{1}{c|}{2067}                   & \multicolumn{1}{c|}{69.3\%}               & \multicolumn{1}{c|}{61.3\%}               & 81.9\%                                       \\ \hline
\multicolumn{1}{l|}{Not direct match} & \multicolumn{1}{c|}{1682}                   & \multicolumn{1}{c|}{64.4\%}               & \multicolumn{1}{c|}{50.9\%}               & 84.0\%                                       \\ \hline
\multicolumn{1}{l|}{Unpopular CUI}    & \multicolumn{1}{c|}{114}                    & \multicolumn{1}{c|}{7.0\%}                & \multicolumn{1}{c|}{2.3\%}                & 14.0\%                                       \\ \hline
\end{tabular}
  \label{tab:analysis} 
\end{table}
\noindent We find that performance is also variable across the UMLS Semantic Types. For some types, the maximum accuracy was nearly perfect: Pharmacologic Substance (93.9\%), Sign or Symptom (94.3\%), Quantitative Concept (95.8\%).  Maximum accuracy was significantly worse for Therapeutic of Preventive Procedures (77.5\%), Medical Devices (78.7\%), and Body Substances (73.3\%). A full table of these results is provided in Appendix B. 

MetaMap and cTAKES underperform compared to competition systems. Over the entire test, accuracy was 59.0\% for MetaMap, 47.1\% for cTAKES, and 64.6\% for both pooled together. This is consistent with past findings in which MetaMap yielded poor performance on the SemEval task for disorder/disease recognition~\citep{Riveros2015}. The poor performance of cTAKES can be partially explained by the fact that its Default Clinical Pipeline is not trained to tag adjectives (e.g. severe, mild). However, both systems are primarily disadvantaged by the fact that they are end-to-end systems. For instance, competition participants were provided with the span \texttt{breast on the left side} which mapped to the CUI for ‘left breast’. In contrast, the end-to-end systems needed to first identify the span themselves, before normalizing them. They only identified the span \texttt{breast}, and not the full span, and therefore tagged with the less specific CUI for ‘breast.’ 

\section{Annotation Framework}
Our analysis clearly found that the performance of clinical normalization systems leaves much room for improvement and that properties of existing corpora make it difficult to evaluate end-to-end normalization. In real pipelines, recognition and normalization are usually not decoupled, and as a result, it's important to be able to evaluate fairly in that setting. Therefore, our goal here was to develop an annotation framework to overcome the limitations of past corpora to allow for a more robust, flexible evaluation. 

\subsection{Framework Development}
The annotation framework was primarily developed by medical residents on our team. We iterated from existing annotation guidelines by annotating notes (from the i2b2/n2c2 dataset and the MIMIC-III Critical Care database from \citet{Johnson2016}), addressing inconsistencies and ambiguities that arose, and noting un-annotated information of particular clinical relevance for downstream systems. Our final framework was grounded both by past issues noted by~\citet{Luo2019} and our own observations.

\subsection{Amended Framework}
We summarize the three largest changes made to existing annotation paradigms changes in Table \ref{tab:schema} below. Additionally, we further explain the motivations behind each of the three significant framework shifts. We attach the full annotation framework in Appendix C.

\begin{table}[htbp]
  \centering 
  \caption{The three significant changes in our annotation schema, in response to the main limitations we found in current annotation frameworks. }
\begin{tabular}{@{}ccc@{}}
\toprule
\textbf{Rule}                                                                                                      & \textbf{Justification}                                                                                                                                                                                   & \textbf{Examples}                                                                                                                                                                                                                                 \\ \midrule
\begin{tabular}[c]{@{}c@{}}Tag multiple\\ CUIs for a\\ single span if \\ they seem\\ equally valid\end{tabular}    & \begin{tabular}[c]{@{}c@{}}We no longer need rules on\\ which CUI to pick. Such rules \\ can be arbitrary and are hard to\\ make universal.\end{tabular}                                                 & \begin{tabular}[c]{@{}c@{}}Sputum $\rightarrow$ Sputum specimen (C0444159)\\     Sputum (C0038056)\\ Stent$\rightarrow$ Stent, device (C0038257)\\                 Vascular stent (C0183521)\end{tabular}                              \\ \midrule
\begin{tabular}[c]{@{}c@{}}Tag \\ subconcepts if\\ they have a \\ medical\\ meaning\end{tabular}                   & \begin{tabular}[c]{@{}c@{}}Concept boundaries are\\ subjective and UMLS is\\ inconsistent, but tagging\\ subconcepts removes ambiguity\\ and reintroduces consistency\\ (e.g. for Figure 3)\end{tabular} & \begin{tabular}[c]{@{}c@{}}Severe Asthma Exacerbation:\\ Severe asthma exacerbation $\rightarrow$ C00038218\\ Severe $\rightarrow$ C0205082\\ Asthma $\rightarrow$ C0004096\\ Asthma exacerbation $\rightarrow$ C0349790\end{tabular} \\ \midrule
\begin{tabular}[c]{@{}c@{}}Tag all medical \\ concepts (e.g.\\ anatomical\\ terms, normal \\ findings)\end{tabular} & \begin{tabular}[c]{@{}c@{}}Problems, tests, and treatments\\  are hard to delineate and are\\  insufficient to understanding\\  a patient's trajectory.\end{tabular}                                     & \begin{tabular}[c]{@{}c@{}}NSR $\rightarrow$ Normal Sinus Rhythm (C0232202)\\ Rheum $\rightarrow$ Rheumatologist (C033489)\\ PEERL $\rightarrow$ Pupils Equal and\\  Reacting to Light (C1261138)\end{tabular}      \\ \bottomrule 
\end{tabular}
  \label{tab:schema} 
\end{table}

\subsubsection{Change 1: Tagging Multiple CUIS due to Term Redundancy} 
The UMLS vocabularies contain redundant terms. For example, there exist separate terms for \textit{Post extubation acute respiratory failure requiring reintubation} and \textit{Acute respiratory failure requiring reintubation}. However, if a patient requires reintubation, they are by definition extubated. The MCN corpus approached this problem by pre-selecting concepts in certain hierarchies to have priority over others. For example, many measurements of substances exist in both ‘Finding’ and ‘Observable Entity’ forms, and the creators of MCN informed annotators to choose the former, if presented with that choice. However, this could have suboptimal effects on model learning, where, for example, representations for the ‘Observable Entity' type are improperly learned. Additionally, such rules are hard to codify over the full domain, so in other cases, an adjudicator over the MCN corpus had to choose one CUI, when the two annotators disagreed. In examining errors from the 2019 n2c2 Shared Task, we did find that many ``errors" arose from models essentially choosing synonyms. Therefore, our framework allows annotators to choose more than one concept, where they find relevant. This also allows for a more accurate evaluation, in addition to cleaner signal during model training. 

\subsubsection{Change 2: Tagging Subspans due to CUI Inconsistencies}
Inconsistencies in available UMLS terms complicate generate universal rules about concept boundaries. 
In the example in Figure 2, we show the concept \texttt{Severe asthma exacerbation} has a term in SNOMED whereas \texttt{severe COPD exacerbation} does not. Additionally existing public clinical entity corpora only tag the CUI corresponding to the longest span of text. Therefore, in the prior example, in one case, \texttt{Severe asthma exacerbation} would be tagged, and in the other only \texttt{COPD exacerbation} would be tagged. Then, if our model trained on \texttt{COPD exacerbation}, our model would likely only tag \texttt{asthma exacerbation}. Under current paradigms, this would count as an incorrect tag. As a result, in our framework, we additionally tag subspans corresponding to valid medical subconcepts, as shown in Table 2. This change would increase the artificially deflated performance numbers from end-to-end systems in our prior analysis, where e.g., they were counted as incorrect for normalizing \texttt{breast} and not \texttt{breast on the left side}.

\begin{figure}[htbp]
  \centering 
  \includegraphics[width=4in]{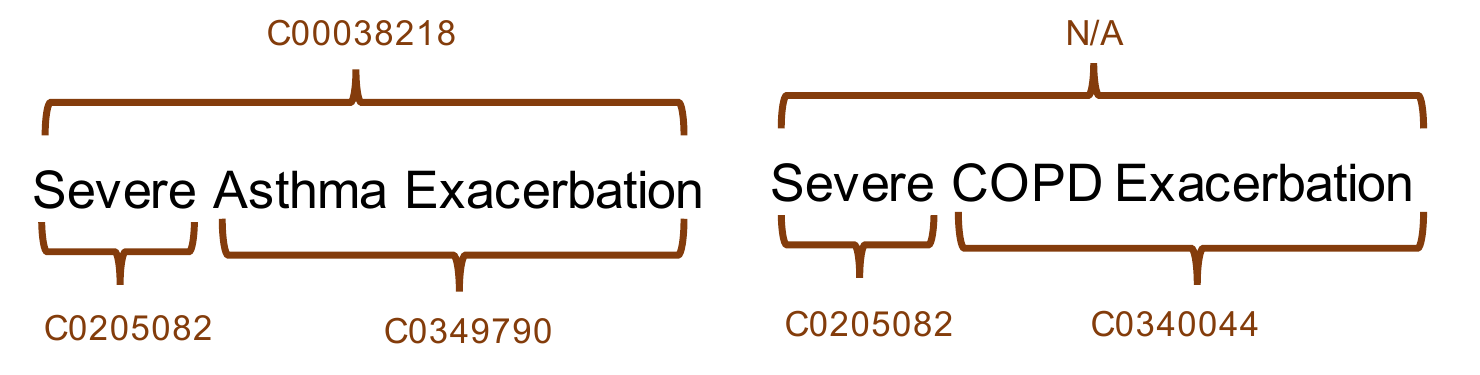} 
  \caption{Examples of compound concepts represented inconsistently in SNOMED. While there is a term for ‘Severe Asthma Exacerbation’, there is no term for ‘Severe COPD Exacerbation.’ Existing datasets only tag the longest concept available in the vocabulary, but this makes it difficult to annotate concepts consistently.}
  \label{fig:example} 
\end{figure}

\subsubsection{Change 3: Increasing Clinical Concept Coverage}
Finally, we found that problems, tests, and treatments are insufficient to describe clinically relevant information. Problems only cover abnormal findings, but normal findings are necessary to place abnormal findings in context. Further, the definition of a problem is dependent on the history of a patient and the degree of the finding. For example, the s3 heart sound is normal in children or athletes, but can be pathologic in older patients. We face similar issues in delineating treatments, when we may also want to capture a lack of intervention. Consider a note that said \textit{Hypoxemia:…on 3-4L NC [nasal cannula] throughout the day yesterday, weaned to RA [room air] this AM}. \texttt{RA [Room air]} isn’t a treatment, but it is clinical terminology that is critical to understanding that the patient’s oxygen requirement changed. 

Finally, outside of issues delineating the three categories, there is other clinically relevant information essential to a patient’s narrative, such as anatomical terms and clinical abbreviations. For example, we may return to the example of \texttt{pt} which in the MCN dataset could refer to prothrombin time, physical therapy, or posterior tibial pulse. In addition, \texttt{pt} is a common abbreviation for patient, and since it is not a problem, test, or treatment, it would never be seen in training; as a result, a downstream end-to-end system would likely face issues. As a result, we believe it is important for datasets to cast a wider net in terms of the clinical terms they tag.

\subsection{Comparison to MCN}
To provide a point of comparison, we compared a note tagged with our schema versus one tagged with the schema from i2b2/MCN. Under our framework, the annotator identified 170 spans with 123 unique CUIs; in MCN, the note had 72 spans with 63 unique CUIs. The changes in our schema manifested in the following ways:
\begin{itemize}
    \item	\texttt{Jaundiced} was tagged both as ‘icterus (C0022346)’ and ‘yellow or jaundiced color (C0474426).’
    \item	The note contained \texttt{place percutaneous drains to decompress his biliary tree}. The span in the MCN corpus was \texttt{percutaneous drains}, which was split into ‘percutaneous (qualifier)’ and ‘drain (device).’ In addition to those, we tagged \texttt{place percutaneous drains} as ‘percutaneous transhepatic insertion of biliary drain (procedure).’ 
    \item Additional annotations included sections of the note (e.g. \texttt{Review of Systems}), normal findings (e.g. \texttt{stable condition}), and medical descriptions (e.g. \texttt{jaundiced}).

\end{itemize}

\section{Validation of Framework}
Next, we ``validate" our framework to show it can yield consistent annotations and that it had the intended effects on evaluation when applied to real world systems. 

\subsection{Annotating under our Framework}
\begin{figure}[htbp]
  \centering 
  \includegraphics[width=4in]{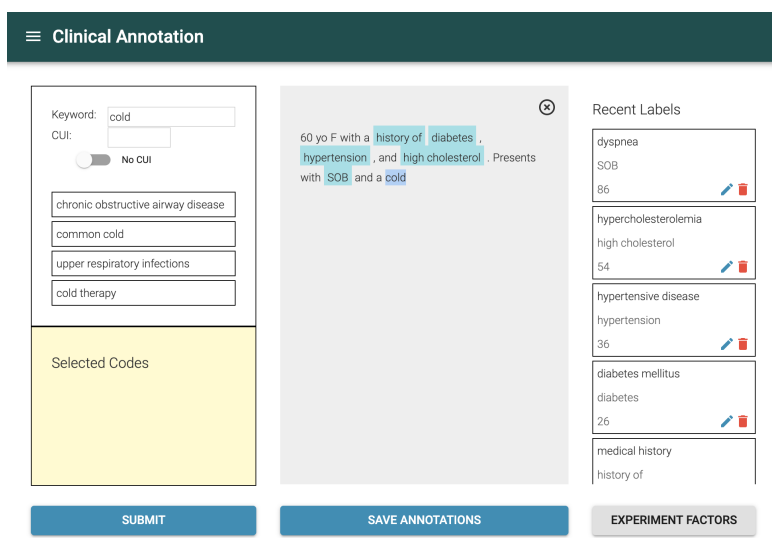} 
  \caption{Custom in-house annotation tool, to be open-sourced. Annotators highlight the span they want to tag in the middle panel (pictured: cold), and the panel on the left automatically provides suggestions for CUIs, which can be scrolled through and selected. If the suggestions provided are insufficient, an annotator can use the UMLS browser and directly enter CUIs instead. The right panel shows the labels that were already tagged and at which character locations: the pencil allows for edits and the trash can for deletions.}
  \label{fig:example} 
\end{figure} 

\noindent Two annotators tagged portions of notes, using the MIMIC-III database, so the resulting dataset could be open-sourced ~\citep{Johnson2016}. They tagged both the History of Present Illness and Assessment and Plan sections of both physician notes and discharge summaries, since these include the bulk of the narrative and the least auto-generated text. 11 note sections were tagged in all. Annotation was conducted using an in-house annotation tool, which is shown and described in detail in Figure 3.

\subsection{Agreement Evaluation}
 We evaluate concordance between the two annotators by measuring span agreement, and for the concordant spans, CUI agreement. A total of 1571 spans were tagged (1288 by Annotator 1, 1435 by Annotator 2). This corresponds to a Jaccard similarity of 0.73 for unadjudicated span selection. Existing clinical entity recognitions datasets have not released agreement numbers for span selection, so we are unable to contextualize this figure; they merely state that they use two annotators, followed by an adjudicator. On concordant spans, annotators matched on at least one CUI 82.6\% of the time (sans adjudication). The MCN dataset reported 67.4\% CUI concordance between annotators pre-adjudication. This is not an apples-to-apples comparison, because by additionally tagging shorter subspans of compound concepts, our system benefits from a smaller proportion of compound concepts. However, we also consider this a strength, since it allows compound concepts to be represented in multiple ways to account for inconsistencies in UMLS.

\subsection{Evaluation of Real World End-to-End Systems}

\begin{table}[ht]
\label{tab:final} 
\begin{tabular}{lc|c|c|c|c|c|c|}
\cline{3-8}
 & \multicolumn{1}{l|}{}                  & \multicolumn{3}{c|}{\textbf{MetaMap}}  & \multicolumn{3}{c|}{\textbf{cTAKES}}  \\ \hline
\multicolumn{1}{|l|}{\textbf{Semantic Type}}                                                                   & \textit{\begin{tabular}[c]{@{}c@{}} \# in \\  Data \end{tabular}} & \textit{\begin{tabular}[c]{@{}c@{}}\% Spans \\ Correct\end{tabular}} & \textit{\begin{tabular}[c]{@{}c@{}}\% CUIs \\ Correct\end{tabular}} & \textit{\begin{tabular}[c]{@{}c@{}}CUI \\ Precision\end{tabular}} & \textit{\begin{tabular}[c]{@{}c@{}}\% Spans\\ Correct\end{tabular}} & \textit{\begin{tabular}[c]{@{}c@{}}\% CUIs\\ Correct\end{tabular}} & \textit{\begin{tabular}[c]{@{}c@{}}CUI\\ Precision\end{tabular}} \\ \hline
\multicolumn{1}{|l|}{\textit{All types}}  & 1271 & 64\%  & 48\%  & 75\% & 52\%  & 46\%   & 89\%  \\ \hline
\multicolumn{1}{|l|}{\textit{Sign or Symptom}}  & 168  & 68\%   & 62\%   & 92\%  & 83\%  & 82\%   & 99\%      \\ \hline
\multicolumn{1}{|l|}{\textit{Disease or Syndrome}}  & 164   & 68\%  & 52\%    & 77\%     & 64\%    & 62\%   & 97\%      \\ \hline
\multicolumn{1}{|l|}{\textit{Finding}}     & 130   & 57\%   & 43\%  & 76\%   & 41\%  & 31\%   & 74\%    \\ \hline
\multicolumn{1}{|l|}{\textit{\begin{tabular}[c]{@{}l@{}}Pharmacologic \\ Substance\end{tabular}}}              & 113   & 80\%     & 75\%   & 94\%    & 86\%  & 84\%    & 98\%   \\ \hline
\multicolumn{1}{|l|}{\textit{Organic Chemical}}    & 102    & 79\%   & 77\%     & 98\%     & 88\%     & 88\%      & 100\%   \\ \hline
\multicolumn{1}{|l|}{\textit{\begin{tabular}[c]{@{}l@{}}Body Part, Organ,\\  or Organ Component\end{tabular}}} & 84  & 40\%    & 22\%   & 55\%            & 72\%   & 66\%     & 92\%   \\ \hline
\multicolumn{1}{|l|}{\textit{\begin{tabular}[c]{@{}l@{}}Therapeutic or\\ Preventive Procedure\end{tabular}}}   & 71    & 58\%   & 30\%    & 51\%    & 38\% & 24\%                     & 63\%   \\ \hline
\multicolumn{1}{|l|}{\textit{Pathologic Function}}   & 53    & 62\%    & 49\%   & 79\%  & 83\% & 68\%  & 82\%    \\ \hline
\multicolumn{1}{|l|}{\textit{Diagnostic Procedure}} & 43  & 70\%  & 47\%   & 67\%  & 51\%   & 44\%  & 86\%  \\ \hline
\multicolumn{1}{|l|}{\textit{Temporal Concept}}  & 41                           & 54\%   & 46\%   & 86\%                                                              & 0\%                                                                 & 0\%                                                                & N/A                                                              \\ \hline

\end{tabular}

\caption{Performance of MetaMap and cTAKES on the whole dataset created under our framework, and subdivided by the ten most common Semantic Types. \textit{\% Spans Correct} indicates what percent of spans were correctly recognized by the model, \textit{\% CUIs correct} indicate what percent of spans were correctly recognized with a correct CUI, and \textit{CUI precision} indicates the percent of correctly identified spans that were normalized correctly.}
\end{table}

Finally, we wanted to use the dataset we created to benchmark MetaMap v4.0.0 (rule-based) and the cTAKES Default Clinical Pipeline (hybrid rule and ML model), since they are two common systems used in the machine learning and healthcare community, and  past comparisons have only focused on narrow concept types. We examine recall on spans in our dataset, since tools like MetaMap also tag terms like \texttt{Date}. We can see results of this analysis in Table 3. For the purpose of the analysis, we considered a union of the annotations created by the annotators, ignoring those spans which did not map to a cUI.

The precision/recall tradeoffs between the two systems are apparent. More interestingly, we find huge discrepancies of accuracies among Semantic Types between the systems. This highlights the utility of benchmarking across a diversity of terms that reflect downstream usage. 

Finally, we also examine the 174 spans that contained at least one subspan and were not contained within any larger span. These are exactly the compound concepts that are inconsistently included in medical vocabularies. Of these, cTAKES only recovered the span in 74 cases. However, of the 100 cases with a missed span, in 54 of them, cTAKES did correctly match a CUI for one of the subspans. In existing schemas where only the longest span is tagged, these 54 (31\%) cases would have all been counted as incorrect, which vastly misrepresents the true performance of cTAKES. We don't see as drastic a bump with MetaMap, since perhaps in contrast to cTAKES, it is entirely rule-based.

\section{Discussion} 
We have shown that performance of entity normalization is far from robust, and accuracy numbers are hugely bolstered by the existence of a few common, easier-to-map terms. As we have shown, concentrating on certain subsets of the data can lead to vastly different metrics. As a result, performance would be highly variable for downstream applications depending on the entity of interest.  This holds true both for the normalization systems from the n2c2 Shared Task and for the widely used MetaMap/cTAKES. Being able to compare systems with a reference dataset is crucial for deciding how to set up a clinical machine learning pipeline.
Our annotation framework could be used to create that dataset, since it allows for more nuanced evaluation. 

In examining the discrepancies between our annotators, we notice they often arose because of minute differences between concepts, e.g. ‘malabsorption’ versus ‘malabsorption syndrome.’ In these cases, note context is insufficient to pick up on subtle differences in definition, which are also generally of little clinical consequence. A system that retrieves either of these should be assessed as correct.

Perhaps most importantly our schema allows for a transition towards end-to-end learning and evaluation. Let us consider an example created from our framework where an annotator tagged \texttt{dirty UA} as ‘urine screening abnormal (Finding)’ and \texttt{UA} as ‘urinalysis (Procedure).’ We can give partial credit to a system that only tags \texttt{UA}. In contrast, under current schemas which only tag the longest concept span, normalizing only to 'urinalysis' would be considered incorrect. Not only does this discount the model's true performance, but it also adversely affects training. Further, in this case, the subspan \texttt{UA} is of a different Semantic Type (Procedure) than the whole span (Finding). If a researcher only wanted to extract procedures, under our framework, they won’t lose out on those spans where procedures are a subspan of a larger concept. Therefore, our schema allows for flexible training of a subset of concepts of interest.



\subsection*{Limitations}


Given that we tag spans in addition to subspans, our annotation framework does greatly increase the number of terms that need to be tagged per document. However, this extra effort is partially offset by the decreased cognitive load of deciding where a span boundary is. Further, by providing suggestions upon highlighting a span, the clinical annotation tool already greatly speeds up the time needed to annotate. Moving forward, we are currently experimenting with having the tool auto-tag unambiguous cases, which the user could simply choose to accept or reject. Since our analysis of n2c2 shows that we can already extract certain terms (e.g. \texttt{hypertension}) and Semantic Types with high confidence, this allows the annotator to focus only on the most difficult 20-30\% of cases. In manual examination of spans tagged only by a single annotator, differences often arose due to annotation fatigue, rather than misunderstanding of the framework. For example, an annotator would not tag a text span they had previously tagged before in the note. This fatigue could be alleviated by the addition of automatic suggestions. 

However, with suggestions and auto-tagging, it is important to understand the sources of bias we could introduce to our resulting dataset. An ongoing area of research is assessing the extent to which the presence, ordering, and presentation of suggestions and auto-tags may bias outcomes. Additionally, we hope to evaluate user interface (UI) techniques to mitigate this bias. There is a need for larger annotated clinical datasets, and smart computational tooling can decrease the time, effort, and cost needed to create them. To that end, we plan to open-source our clinical annotation tool, as well as the annotations generated by this work. See \texttt{https://github.com/clinicalml/mimic\_annotations}. 

We do acknowledge that the annotation guidelines were developed and evaluated using notes written in a critical care context, so that we could open-source the resulting MIMIC data set. Future work remains to evaluate the schema in the context of other clinical settings and note types.

\section{Conclusion}
Medical notes offer us an incredibly rich view into patient narratives, but they have remained a largely untapped resource for clinical and machine learning research. Clinical studies often resort to manual chart review, which is time-consuming and difficult to scale. While automated concept extraction systems provide a way to exploit clinical text, as we have demonstrated in this work, they cannot yet be reliably deployed.

Developing truly robust and accurate clinical entity recognition and normalization algorithms will require both substantially more labeled training data and automated evaluation metrics that account for the subtlety and ambiguity of the task. We present a new annotation framework and a small-scale annotated dataset for evaluation of end-to-end concept recognition and normalization. The time is ripe for the field to invest in substantially larger labeled training sets to spur new machine learning approaches for clinical entity recognition and normalization. Future work should consider incorporating the new annotation framework as the basis for such a data set.

\acks{The authors thank Irene Chen and Divya Gopinath for their helpful comments on the manuscript, and thank Noemie Elhadad, Anna Rumshisky, and Ozlem Uzuner for their valuable conversations and perspective on the clinical annotation process. They would also like to the thank all the organizers of the 2019 n2c2 Workshop for sharing the outputs of top systems for analysis.}

\bibliography{ref}
\newpage
\appendix
\section*{Appendix}

\subsection*{Appendix A: Description of Annotation Tool Suggestion System}

Below, we describe the lightweight suggestion system that underlies our annotation tool. Due to its simplicity, it can run seamlessly. 
First, we constructed two data structures from UMLS tables, using only the terms from SNOMED and RxNorm. 
\begin{itemize}
    \item A lookup table. This is a dictionary mapping text to all of the CUIs that have that text as a Preferred Name or Synonym. 
    \item An inverted index. The inverted index was built as per classical information retrieval. The ‘document’ for a CUI was a concatenation of stemmed versions of all the words in that CUI’s Preferred Name or Synonym. 
\end{itemize}
The suggestions then consist of: 
\begin{itemize}
    \item Any direct matches of the text span, found in the lookup. They are sorted by the number of synonyms they have, which we use as a proxy for population prevalence.
    \item The closest set of matches from the inverted index. Before searching, we remove stop words from the highlighted text and stem the remainder of the words.  Among equally relevant matches, we again sort by number of synonyms as a proxy for population prevalence.
\end{itemize}
\newpage
\subsection*{Appendix B: Performance of Systems by Semantic Type}
Below we show the performance of the top 10 systems of the n2c2 Shared Task on the test set, subdivided by the 20 most common Semantic Types. We note that a CUI can have multiple Semantic Types, in which case it is included in several calculations. \\

\begin{adjustbox}{center}
\begin{tabular}{lcccc}

\textbf{Test Subset}                                                            & \textbf{\# of Examples} & \textbf{Max Accuracy} & \textbf{Avg Accuracy} & \textbf{Pooled Accuracy} \\ \hline
Disease or Syndrome                                                             & 713                     & 88.8\%                & 85.7\%                & 96.2\%                   \\ \hline
Finding                                                                         & 674                     & 78.8\%                & 75.1\%                & 86.4\%                   \\ \hline
Pharmacologic Substance                                                         & 667                     & 93.9\%                & 91.9\%                & 97.2\%                   \\ \hline
Laboratory Procedure                                                            & 665                     & 88.7\%                & 85.8\%                & 94.4\%                   \\ \hline
Organic Chemical                                                                & 624                     & 95.5\%                & 94.4\%                & 98.1\%                   \\ \hline
\begin{tabular}[c]{@{}l@{}}Therapeutic or\\ Preventative Procedure\end{tabular} & 565                     & 77.5\%                & 72.7\%                & 89.4\%                   \\ \hline
Sign or Symptom                                                                 & 456                     & 94.3\%                & 92.3\%                & 96.9\%                   \\ \hline
Diagnostic Procedure                                                            & 411                     & 90.3\%                & 84.6\%                & 95.4\%                   \\ \hline
Qualitative Concept                                                             & 311                     & 94.9\%                & 89.0\%                & 96.8\%                   \\ \hline
Health Care Activity                                                            & 299                     & 84.6\%                & 81.6\%                & 89.3\%                   \\ \hline
Spatial Concept                                                                 & 241                     & 92.5\%                & 89.0\%                & 95.4\%                   \\ \hline
Pathologic Function                                                             & 228                     & 89.0\%                & 85.8\%                & 94.7\%                   \\ \hline
\begin{tabular}[c]{@{}l@{}}Body Part, Organ,\\  or Organ Component\end{tabular} & 189                     & 82.0\%                & 65.2\%                & 89.9\%                   \\ \hline
Quantitative Concept                                                            & 165                     & 95.8\%                & 92.8\%                & 97.0\%                   \\ \hline
Temporal Concept                                                                & 122                     & 85.2\%                & 83.6\%                & 91.8\%                   \\ \hline
Medical Device                                                                  & 108                     & 78.7\%                & 70.0\%                & 88.0\%                   \\ \hline
Antibiotic                                                                      & 102                     & 97.1\%                & 95.6\%                & 98.0\%                   \\ \hline
Neoplastic Process                                                              & 100                     & 91.0\%                & 84.8\%                & 97.0\%                   \\ \hline
Functional Concept                                                              & 93                      & 87.1\%                & 77.0\%                & 88.2\%                   \\ \hline
\begin{tabular}[c]{@{}l@{}}Amino Acid, Peptide,\\ or Protein\end{tabular}       & 71                      & 91.5\%                & 86.9\%                & 91.5\%                  
\end{tabular}
\end{adjustbox}

\newpage 
\subsection*{Appendix C: Full Clinical Annotation Schema Guidelines}
\textit{\large Overall Guidelines} \\
Tag all words that seem like medical terms, i.e. they require medical knowledge to understand or are frequently used in medicine.\\

\noindent \textit{\large Concepts and Actions (Nouns and Verbs)} 
\begin{itemize}
\item	Tag all labs, e.g. "WBC count", "hematocrit", or "urinalysis".
\item	Tag all procedures, e.g. "EKG", "chest x-ray", "fingerstick", or "CT".
\item	Tag all treatments or interventions, e.g. "transfusion", "aspirin", "CABG", or "surgery".
\begin{itemize}
    \item Tag interventions even if they’re common non-medical concepts, e.g. "fluids" or "tube".
    \item	Tag normal English verbs that are used to describe an intervention, e.g. for “blood cultures were drawn” tag “drawn” as “sample obtained”.
    \item	For drugs, tag the drug name and dosage route, but do not tag the dosage. E.g. do tag “PO” but do not tag “15mg”.
\end{itemize}
\item	Tag all symptoms, e.g. "headache" or "nausea".
\begin{itemize}
\item	Tag symptoms even if they are not medical terms, i.e. for “chest pain” tag “chest”, “pain”, and “chest pain”.
\end{itemize}

\item	Tag all findings, e.g. "respiratory distress", "edema", "blood pressure", or "heart rate".
\item	Tag all diagnoses, e.g. "COPD", "MI", or "osteoporosis".
\item	Tag all medical history elements, e.g. "trauma" or "smoking history".
\item	Tag all healthcare terms, e.g. "PCP", "ED", "ambulance", "EMR", "OSH", "cardiologist", "rheum" (or any other medical specialty), "follow-up", "review of systems", "past medical history", "patient", "assessment", or "consult".
\item	Tag all illnesses or medical findings, e.g. "injury".
\item	Do not tag negation words, e.g. for “denies chest pain” do not tag “denies”.\\
\end{itemize} 
\noindent \textit{\large Modifiers (Adjectives and Adverbs)}
\begin{itemize}
\item Tag all words that modify a medical concept.
\item	Tag all medical modifiers, e.g. "acute", "pleuritic", or "friable".
\item	Tag all anatomical terms, any words that refer to body positioning, e.g. "orbital", "right", or for “chest x-ray” tag “chest”.
\item	Tag all descriptions of findings or a patient’s state, e.g. "febrile" or "stable".
\begin{itemize}
\item	Do not tag adjectives that describe something medical but do not have a separate medical meaning from their English meaning, e.g. for “vital signs normal” do not tag “normal”.
\end{itemize}
\item	Tag all descriptions of diseases or diagnoses that are considered medical terms or have separate medical meanings, e.g. "acute", "chronic", "mild", "moderate", "severe", "dynamic".
\begin{itemize}
    \item For example, tag “sensitive” if it is in the context of antimicrobial sensitivity, e.g. for “blood cultures grew E. Coli sensitive to x, y, z” tag “sensitive” because it describes a medical concept of antimicrobial sensitivity.
\item 	Do not tag a term if you search for it and there’s no tag available, e.g. "profuse".
\item	Do not tag the words "likely", "unlikely", or "possible". \end{itemize}
\end{itemize}

\textit{\large References to Parts of the Note}
\begin{itemize}
    \item 	Tag references to parts of notes even though they are English words, but do not tag individual words that comprise them.
\begin{itemize}
\item	For example, "HPI" (History of Present Illness), "PMH" (Past Medical History),"A/P" (Assessment and Plan) \\
\end{itemize}

\end{itemize}

\textit{\large Span Splitting}
\begin{itemize}
    \item	Tag sub-concepts if they have their own medical meaning, e.g. for “extraocular movements intact” tag “extraocular”, “extraocular movements”, and “extraocular movements intact”. However, in the case of "respiratory failure" the word "failure" does not have any separate medical, anatomical, or physiological relevance so it would not be tagged.  
\item	Do not tag sub-concepts if they are contained within other concepts that express the same meaning more fully written out, e.g. for “Crohn’s disease” tag “Crohn’s disease” and “disease” but not “Crohn’s”.
\item	Do not tag sub-spans of very common administrative terms used to describe portions of a note, e.g. for “history of present illness” do not tag “history”, “present”, or “illness”. \\
\end{itemize}

\textit{\large Multiple Concept Matches }
\begin{itemize}
\item	Tag multiple CUIs when there are multiple that appear to be a good match in context, e.g. for “fever” potentially tag “fever symptom (finding)” and “fever (sign/symptom)”.
\item	Even if a term can have multiple taggable meanings, only tag the meaning in context.
\item	If two terms match equally well, only tag the verbatim text, e.g. for “acute kidney injury” only use tag “acute kidney injury” but not “kidney failure, acute”.\\
\end{itemize}

\textit{\large No Concept Matches}
\begin{itemize}
\item 	If there is no exact match in SNOMED, tag as “CUI-less”  (and optionally any close approximations)
\begin{itemize}
    \item If both subconcepts in a compound concept are tagged, there is no need to tag the compound concept as "CUI-less."
    \item Do not use “CUI-less” to tag terms that are only semantically off (the concept "patients" for "patient"). 
 
\end{itemize}
\item If there is an approximate match, feel free to tag as both "CUI-less" and those close approximations, e.g. "presentation" for "presents with".\\
\end{itemize}

\textit{\large Special Considerations}
\begin{itemize}
    \item If a term repeats within the same note, tag the term every time it appears.
\item Ignore typos and tag the term as its intended meaning.
\item Tag common medical abbreviations, e.g. "h/o", "PMH", "ggt", or "p/w".
\begin{itemize}
    \item 	Do not split abbreviations into different spans, even if they represent different medical concepts, e.g. for “EOMI” do not tag “EO” and “EOM” separately.
\end{itemize}
\item For simplicity, tag lab tests with priority order level (aka finding) $\textgreater$ measurement (procedure) $\textgreater$ substance, e.g. for “albumin level” tag with “albumin level measurement” instead of “albumin”. While multiple may be equally valid, this can be fixed in post-processing.
\begin{itemize}
    \item 
	If a phrase describes the results of a lab test, only tag the substance tested itself and not the modifier, e.g. for “low albumin” tag “albumin” and do not tag “low”.
\item	Similarly, do not tag the numeric value if given, e.g. for “INR 3.1” do not tag “3.1”.
\item	Do tag lab test results with modifiers if the resulting state has a name, e.g. for “elevated WBC count” tag the phrase as “leukocytosis”.
\end{itemize}
\item 	Do not tag units of measurement.
\end{itemize}

\end{document}